%% file: main.tex
\pdfoutput=1

\documentclass[11pt]{article}

\usepackage{acl}

\input{sections/math_commands.tex}

\usepackage{times}
\usepackage{latexsym}

\usepackage[T1]{fontenc}

\usepackage[utf8]{inputenc}

\usepackage{microtype}

\usepackage{inconsolata}

\usepackage{graphicx}
\usepackage{tikz}
\usepackage{pgfplots}

\usepackage{tabularx, booktabs}
\newcolumntype{Y}{>{\centering\arraybackslash}X}
\usepackage{multirow}

\usepackage{xcolor}

\usepackage{subcaption}

\usepackage[ruled]{algorithm2e}
\SetKwComment{Comment}{$\triangleright$\ }{}
\SetAlFnt{\small}

\usepackage{float}

\title{Hybrid-RACA: Hybrid Retrieval-Augmented Composition Assistance \\ for Real-time Text Prediction}

\author{Menglin Xia$^*$ \qquad Xuchao Zhang$^*$ \qquad Camille Couturier \\
{\bf Guoqing Zheng} \qquad {\bf Saravan Rajmohan} \qquad {\bf Victor R\"uhle} \\
Microsoft \\
\{mollyxia, xuchaozhang, cacoutur, zheng, saravar, viruh\}@microsoft.com
  }

\begin{document}
\maketitle
\def\thefootnote{*}\footnotetext{These authors contributed equally to this work.}\def\thefootnote{\arabic{footnote}}

\begin{abstract}
Large language models (LLMs) enhanced with retrieval augmentation has shown great performance in many applications. However, the computational demands for these models pose a challenge when applying them to real-time tasks, such as composition assistance. To address this, we propose Hybrid Retrieval-Augmented Composition Assistance (Hybrid-RACA), a novel system for real-time text prediction that efficiently combines a cloud-based LLM with a smaller client-side model through retrieval augmented memory. This integration enables the client model to generate better responses, benefiting from the LLM's capabilities and cloud-based data. Meanwhile, via a novel \textit{asynchronous} memory update mechanism, the client model can deliver real-time completions to user inputs without the need to wait for responses from the cloud. Our experiments on five datasets demonstrate that Hybrid-RACA offers strong performance while maintaining low latency.
\end{abstract}

\input{sections/1_introduction}
\input{sections/2_related_work}

\input{sections/3_model}

\input{sections/4_experiment}

\input{sections/5_conclusion}

\section*{Broader Impact}
In our research, we present a pioneering approach to the future landscape of AI applications, envisioning a hybrid system that brings the best of client and cloud worlds. Our unique design allows client and cloud models to function seamlessly in a composition assistance scenario, achieving better performance by levering cloud models and data, and ensuring low-latency and cost-effectiveness by utilizing on-device client models. We believe that our hybrid solution with asynchronous communication is also a valuable solution to make advanced AI more accessible to a wider range of users, including those in resource-constrained environments or with limited access to high-speed internet connections. We believe that our vision can be extended to more applications not limited to composition assistance. Furthermore, our efficient solution, which combines edge and cloud computing, offers great potential to energy conservation. By minimizing the necessity to access resource-intensive large language models (LLMs), notorious for their high energy consumption, our approach mitigates potential harm to the environment. This not only underscores our commitment to sustainability but also highlights the practical benefits of our technology in addressing energy challenges.

\section*{Ethical Considerations}
Hybrid-RACA is a composition assistance tool that integrates client and cloud models and data. In our implementation, data is transmitted between the client and cloud as plain text. However, this transmission process poses potential privacy and confidentiality risks. To mitigate these risks, security measures such as cryptography and access controls can be implemented. When instruction-tuning the client model, we used LLMs to generate completions, which can be considered as a form of synthetic data generation. Like other work that leverages LLMs, this might raise privacy and copyright concerns. We are committed to follow the best practices currently available to minimize privacy and copyright risks by conducting experiments on public datasets and adopting security guardrails.


\bibliography{main}

\appendix

\input{sections/6_appendix}

\end{document}

%% file: sections/math_commands.tex
\usepackage{amsmath,amsfonts,bm}

\def\eqref#1{equation~\ref{#1}}

\def\1{\bm{1}}

\def\vx{{\bm{x}}}
\def\vy{{\bm{y}}}

\DeclareMathAlphabet{\mathsfit}{\encodingdefault}{\sfdefault}{m}{sl}
\SetMathAlphabet{\mathsfit}{bold}{\encodingdefault}{\sfdefault}{bx}{n}



%% file: sections/1_introduction.tex
\section{Introduction}


Large language models have become powerful tools in language processing and they are widely adopted across applications. When augmented with retrieved documents \citep{lewis2020retrieval, liu2022knowledge}, these models can generate more relevant and useful responses. However, the large size of these models and the additional retrieval step introduce significant computational overhead. This leads to increased latency and higher operational costs, limiting their effectiveness in real-time applications, such as composition assistance.

Real-time \textit{composition assistance} tools are designed to swiftly suggest next words or sentences to help users write faster. These systems must operate within tight latency budgets, and they are frequently triggered as the user types. To minimize latency (including model inference latency and communication to the cloud) and to reduce costs, these models are usually deployed on users' edge devices. This imposes strict constraints on the model's size and capabilities, limiting the effectiveness of composition assistance. While recent advancements have enabled models such as Llama \cite{touvron2023llama} to run on smaller devices\footnote{\url{https://github.com/ggerganov/llama.cpp}}, they still fall short in terms of achieving real-time responses. 

For real-time tasks, we encounter a dilemma: LLMs offer superior performance but they are slow and expensive to run, whereas client models are agile and efficient but limited in performance. 
Hybrid computing between client and cloud models is a promising approach to bridge the gap between the challenges of latency and model performance. However, in existing hybrid computing patterns, such as model routing and split computing \cite{kudugunta-etal-2021-beyond-distillation, matsubara2022split}, client and cloud models usually function with synchronized communication. This means that whenever the cloud model is utilized, the system must wait for the cloud model to complete its processing before producing the output. Therefore, simply applying existing hybrid patterns to cloud-based LLMs will not resolve the issue of latency and cost. Besides, existing hybrid patterns usually overlook cloud-based data, which could be essential for effective composition assistance, such as accessing relevant documents in companies' cloud storage.

\begin{figure*}[!t]
    \centering
    \includegraphics[width=\textwidth, trim={1.0cm 0.1cm 0.3cm 0.0cm}]
    {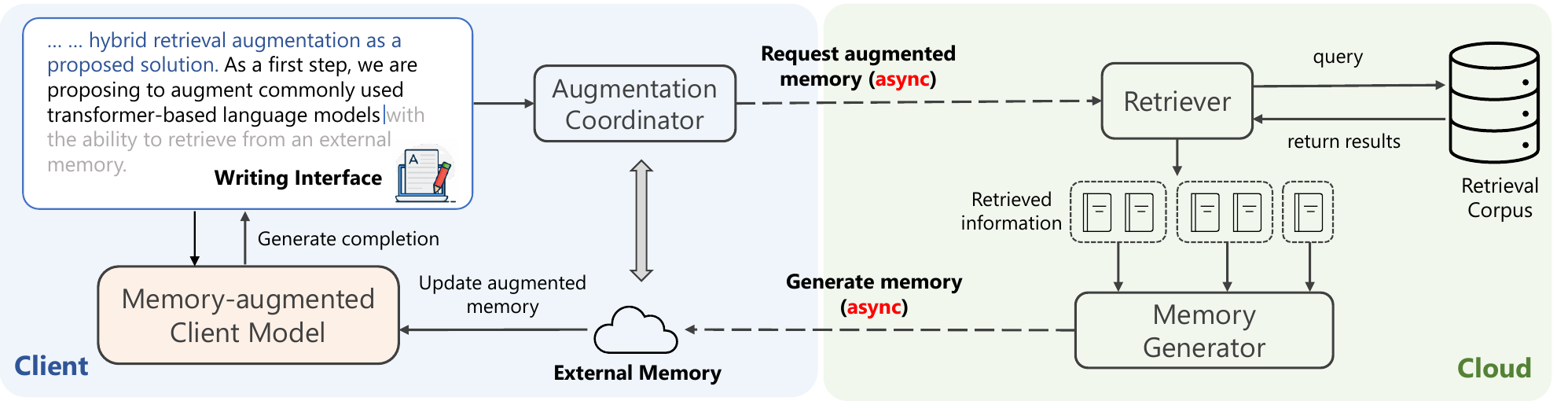}
    \caption{Overview of the Hybrid-RACA system, which is a hybrid system for composition assistance. The top left box represents the writing interface. The framework has four main components: augmentation coordinator and client model on the client side (left), and retriever and LLM-based memory generator on the cloud (right).}
    \label{fig:overview}
\end{figure*}

To address these challenges, we propose a novel Hybrid Retrieval-Augmented Composition Assistance (Hybrid-RACA) system (see Figure \ref{fig:overview}). This system leverages a cloud LLM and cloud data to boost the performance of small language models on client devices through retrieval augmentation, while operating \textit{in an asynchronous manner}. Hybrid-RACA consists of an augmentation coordinator and a small model for text prediction deployed on client devices, as well as a retriever and an LLM located on the cloud server. The client augmentation coordinator sends asynchronous request to the cloud. The cloud retrieves relevant documents and employs an LLM to compress the retrieved documents into shorter snippets of information, which we refer to as \textit{memory}, and sends it asynchronously to the client. On the client side, an instruction-tuned client model leverages available memory to suggest the next words.

The Hybrid-RACA system offers several benefits. (1) \textit{Enhanced utility}: Hybrid retrieval augmentation enables the client model to make better suggestions by leveraging cloud-based resources. (2) \textit{Low latency}: Asynchronous augmentation allows the client to make predictions without waiting for the cloud. This mitigates the effects of network latency and avoids slow inference inherent to cloud-based retrieval-augmented LLMs. (3) \textit{Reduced client-to-cloud communication}: the augmentation coordinator minimizes the client-to-cloud communication by requesting augmented memory only when existing memory becomes stale, reducing the frequency of calling the cloud models and thus saving cost. Furthermore, using LLM-compressed memory further reduces data transfer volume. 

To evaluate our system, we conduct experiments on the text prediction task on five datasets from diverse domains. We compare our model to several baselines and show that our model exhibits substantial utility improvement in text prediction while maintaining low latency. The code for our system will be made available at: \url{https://github.com/microsoft/hybrid-raca}. 

%% file: sections/2_related_work.tex
\section{Related Work}\label{sec:related_work}

\paragraph{Hybrid Computing} 

Hybrid computing divides processing tasks between the edge and the cloud, effectively addressing the limited computation capabilities of edge devices and enabling real-time responses of critical services \citep{loghin2019towards, wang2020asurvey}. For example, split computing partitions machine learning modules between edge and cloud devices to balance overall computation cost and efficiency \citep{matsubara2022split, osia2020ahybrid}. Communication between edge and cloud in split computing is inherently synchronized, as both devices contribute to completing one inference run. More recently, task-specific model routing \cite{kudugunta-etal-2021-beyond-distillation} has also emerged as a promising approach for hybrid computing via routing between client and cloud models. Nonetheless, the overall system still needs to wait for the cloud model whenever it is used, thus limiting the overall latency. 
Another notable paradigm for hybrid computing in machine learning is federated learning, which leverages multiple computing devices for training machine learning models for safety or efficiency purposes \citep{bonawitz2019towards}. However, this technique is less commonly used for inference. 
In addition to hybrid computing, there is also literature on improving efficiency of models deployed on edge devices \citep{tambe2021edgebert} as well as methods on reducing the size of large models for deployment on smaller devices \citep{hoefler2021sparsity}. These methods are orthogonal to our work.

\paragraph{Retrieval Augmented Models}

Retrieval augmentation enhances a language model with retrieved information from external databases. Various methods have been proposed to integrate the retrieved data into the language model, including the use of prompts \citep{lewis2020retrieval, guu2020realm, shi2023replug}, cross-attention modules \citep{borgeaud2021retro}, vector concatenation \citep{izacard-grave-2021-leveraging, fan-etal-2021-augmenting}, and output distribution adjustment at decoding \citep{Khandelwal2020Generalization, liu2022knowledge}. In this work, we adopt the prompting method, which incorporates retrieved data into the input. 
However, the Hybrid-RACA system can be extended to other retrieval augmentation approaches.

%% file: sections/3_model.tex
\section{Hybrid-RACA}

We present our Hybrid-RACA system that leverages cloud-generated memory to enhance the utility of client-based language model while maintaining low latency for composition assistance.

In Hybrid-RACA, the augmentation coordinator (client) monitors the writing context and sends an asynchronous request for an augmented memory from the cloud. The retriever on the cloud searches for relevant data upon request. Subsequently, The memory generator (cloud) leverages an LLM to construct a memory that includes all essential information from the retrieved data, optimizing its usefulness. Finally, the memory is transmitted to the client and seamlessly integrated into the client model for offering real-time suggestions. Algorithm \ref{alg:workflow} describes the inference workflow of Hybrid-RACA.

In the following subsections, we discuss the details of the four main components.

\subsection{Augmentation Coordinator}\label{sec:coordinator}
\begin{figure}[t]
    \centering
    \includegraphics[width=0.45\textwidth, trim={0.8cm 0.5cm 0.3cm 0.0cm}]
    {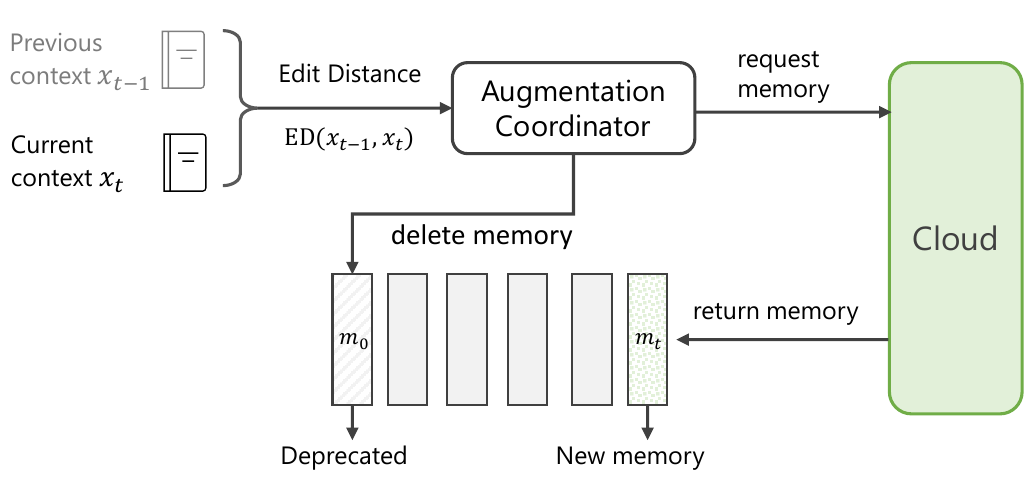}
    \caption{Process of the augmentation coordinator}
    \label{fig:coordinator}
\end{figure}

The augmentation coordinator manages the augmented memory $\mathcal{M}$ by monitoring changes to the \textit{writing context}, which we define as the text the user has already typed (see Fig.\ref{fig:coordinator}). To determine whether a memory update is necessary, the coordinator takes into account the current context $\vx_{t}$ and the context $\vx_{t-1}$ from the previous step and calculates the edit distance $\mathrm{ED}(\vx_t, \vx_{t-1})$. Once the distance exceeds a pre-determined threshold $\tau$, the coordinator initiates a request to the cloud server asking for a new memory. We employ the Levenshtein distance \citep{yujian2007normalized} to measure token-level difference. To avoid redundant memory requests, we adopt an incremental memory update approach, where only the newly updated context is used as the query input to generate the new memory $m_t$. 
When the augmented memory reaches its maximum capacity of $\mathcal{M}$, the oldest memory $m_0$ is deprecated and replaced by the new memory $m_t$.



\begin{algorithm*}[t]
\footnotesize
\caption{Inference workflow of Hybrid-RACA}
\label{alg:workflow}
\KwData {current user input $\vx_t$, input history $\vx_{t-1}$, retrieval corpus $\mathcal{D}$, retrieval model $\mathcal{M}_\mathrm{retrieval}$, cloud-based LLM $\mathcal{M}_\mathrm{cloud}$, client model $\mathcal{M}_\mathrm{client}$, memory $\mathcal{M}$}

\While{$\vx_t$}{
$\mathrm{ED_t} = \mathrm{Edit Distance}(\vx_t, \vx_{t-1})$ \Comment*[r]{Compute changes in context}
\If{$\mathrm{ED_t} > \tau $ \Comment*[r]{Send async request to the cloud}}{
\textbf{async} $\mathcal{D}_r = \{d_1, ... d_k\}$: $\mathcal{D}_r \sim \mathcal{M}_\mathrm{retrieval}(\vx_t, \mathcal{D})$ \Comment*[r]{Retrieve relevant documents}
\textbf{async} $m_t \sim \mathcal{M}_\mathrm{cloud}(\mathcal{D}_r)$ \Comment*[r]{Generate memory}
$\mathcal{M} = Update(\mathcal{M}, m_t)$ \Comment*[r]{Update $\mathcal{M}$ with $m_t$}
Sample $\vy_t \sim \mathcal{M}_\mathrm{client}(\vx_t, \mathcal{M})$ \Comment*[r]{Text prediction with the client model}
\eIf{$Accept(\vy_t)$}{
$\vx_{t-1} \gets \{\vx_{t-1}, \vx_t\}, \vx_t \gets \{\vx_t, \vy_t\}$ \Comment*[r]{User accepts suggestion}
}
{
$\vx_t \gets \{\vx_t, Input()\}$ \Comment*[r]{User rejects suggestion and enters new input}
}
}}
\end{algorithm*}

\subsection{Retrieval-Augmented Memory Generator} \label{sec:mem_generator}

Upon receiving a request from the augmentation coordinator, the memory generator on the cloud initiates the preparation of the augmented memory, which will be returned to the client. The memory preparation process consists of two steps: document retrieval and memory generation.

\paragraph{Document Retrieval}

Given an input query $\vx$, the goal of the retriever is to retrieve the most relevant documents $\mathcal{D}_r=\{d_1, \dots, d_k\}$ from a large corpus $\mathcal{D}$, where $\mathcal{D}_r \subseteq \mathcal{D}$. We use the Dense Passage Retrieval (DPR) \citep{karpukhin2020dense} model in our implementation. 

\paragraph{Memory Generation}

After retrieving the relevant documents $\mathcal{D}_r$, we employ a LLM to generate concise key takeaways that capture essential information from the documents. We use the key takeaways instead of the original documents because the client model is a small language model that usually struggles with processing long context and has a strict limit on input context length. Additionally, extracting key takeaways significantly reduces the memory size, resulting in lower communication and inference cost for the client. 

To generate key takeaways from retrieved documents $\mathcal{D}_r$, we split the documents into chunks and utilize an LLM to extract key takeaways from each chunk. To minimize the frequency of LLM calls, we consolidate multiple chunks within one document.
Subsequently, all generated key takeaways from retrieval documents are merged to form the memory $m_t$ for the current $t$-th memory request. 

\subsection{Memory-Augmented Client Model} \label{sec:client_model}


			
	
	



\begin{figure*}[!t]
    \centering
    \includegraphics[width=0.9\textwidth]
    {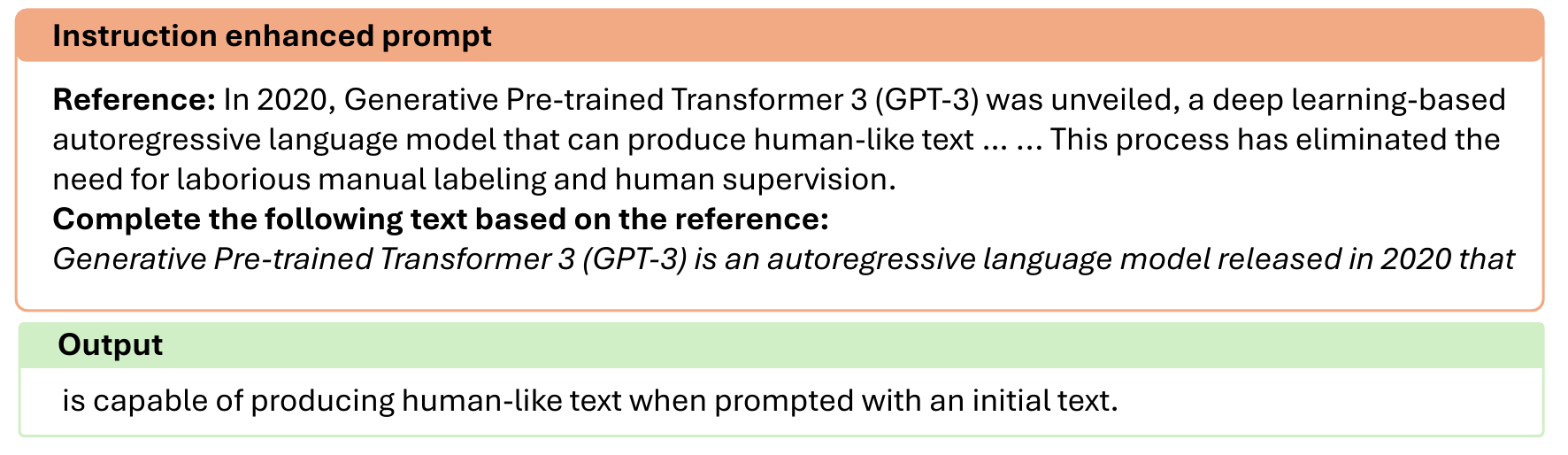}
    \caption{Example of constructing instruction-tuning data}
    \label{fig:instruction}
\end{figure*}

The goal of the client model is to generate useful completions to the user input. Enhanced by cloud-generated memory, our client model learns to make more relevant predictions. We further adopt instruction-tuning to bolster the client model's ability to effectively leverage cloud-generated memory. 

To instruction-tune the client model, we leverage an LLM to generate the instruction tuning data. Given a document $d$, we use the beginning part of the document as the input prompt $\vx=\mathcal{I}(d)$ and use $\vx$ to generate the augmented memory $\mathcal{M}$. We formulate an instruction-enhanced prompt to instruct the model to make predictions based on the memory (see Fig.\ref{fig:instruction}). As for the ground truth labels $\hat{\vy}$, a straightforward approach is to directly use the remaining part of the document $d$. However, this is not ideal as there is usually a discrepancy between the original text and the memory, which can negatively impact the performance of the client model. To address this, we employ an LLM to generate the labels $\hat{\vy} = \mathcal{M}_\mathrm{cloud}(\mathcal{I}(d), \mathcal{M})$.

Then we finetune the client model on the instruction-enhanced prompt and the LLM-generated labels. The model is finetuned on the task to predict $\hat{\vy}$ given $\vx$ and $\mathcal{M}$.  To minimize the discrepancy between our model's predictions $\vy$ and the LLM-generated labels $\hat{\vy}$, we employ the cross-entropy loss on the generated tokens in finetuning:
\begin{equation} \label{eq:loss}
    \mathcal{L}_d=-\sum_{i=1}^l \hat{y_i} \log\bigg(p_\theta(y_i|\vx, \mathcal{M}, \hat{y}_{<i})\bigg)
\end{equation}
where $l$ is the length of the label and $p_\theta(\cdot)$ refers to the probability of tokens generated by the client model.

%% file: sections/4_experiment.tex
\section{Experiments}\label{sec:experiments}
In this section, we introduce the experimental setup (Section \ref{sec:experiment_setup}) and present the evaluation results of Hybrid-RACA system on utility (Section \ref{sec:utility}), inference latency (Section \ref{sec:latency}) and effects of asynchronous memory update (Section \ref{sec:async}).

\begin{table*}[t!]
\centering
\begin{small}
    
    \begin{tabularx}{1.0\textwidth}{c|c|c *{6}{Y}}
    \toprule
     &   &  PPL  &  GLEU  &  BLEU4  &  ROUGE1  &  ROUGEL  &  METEOR  &  BERTScore  \\
    \midrule
    \multirow{4}{2.5em}{OPT-125M}
     &  Vanilla OPT	       & 9.3 & 11.4 & 6.9 & 27.5 & 22.1 & 20.2 & 84.0 \\
     &  HybridRAG	               & 4.3 & 12.8 & 9.6 & 28.4 & 23.4 & 22.4 & 84.5 \\
     &  Hybrid-RACA w/o FT	   & 3.8 & 14.7 & 12.2 & 29.9 & 25.1 & 24.3 & 84.8 \\
     &  Hybrid-RACA FT        & 3.4 & 23.0 & 21.4 & 39.6 & 32.8 & 34.4 & 87.0 \\
     &  Hybrid-RACA IT        & \textbf{2.6} & \textbf{30.2} & \textbf{28.8} & \textbf{48.3} & \textbf{40.2} & \textbf{44.1} & \textbf{89.0} \\
    \midrule
    \multirow{4}{2.5em}{OPT-350M}
     &  Vanilla OPT	       & 7.4 & 13.2 & 8.8 & 30.1 & 24.3 & 22.8 & 84.8 \\
     &  HybridRAG	               & 3.6 & 15.4 & 12.5 & 31.6 & 26.0 & 25.6 & 85.4 \\
     &  Hybrid-RACA w/o FT	   & 3.3 & 17.6 & 15.4 & 33.5 & 27.9 & 28.0 & 85.7 \\
     &  Hybrid-RACA FT        & 3.2 & 23.9 & 22.3 & 40.7 & 33.8 & 35.5 & 87.4 \\
     &  Hybrid-RACA IT        & \textbf{2.4} & \textbf{32.6} & \textbf{31.4} & \textbf{50.8} & \textbf{42.9} & \textbf{46.6} & \textbf{89.5} \\
    \bottomrule
    \end{tabularx}
\end{small}
\caption{In-domain evaluation of Hybrid-RACA performance}
\label{table:utility_wikitext}

\end{table*}

\begin{table*}[t!]

\centering
\begin{small}
	\begin{tabularx}{1.0\textwidth}{cc|c *{1}{Y}|c *{1}{Y}|c *{1}{Y}|c *{1}{Y}}
		    
			\toprule
			& & \multicolumn{2}{c}{Enron Emails} & \multicolumn{2}{c}{NIH ExPorter} & \multicolumn{2}{c}{Hacker News} & \multicolumn{2}{c}{Youtube Subtitles}\\
			\cmidrule(lr){3-10} 
			& & PPL & GLEU & PPL & GLEU & PPL & GLEU & PPL & GLEU\\
            \midrule
            \multirow{4}{2.5em}{OPT-125M}

			& Vanilla OPT	       & 8.5 & 5.8 & 7.4 & 9.3 & 7.5 & 8.0 & 9.2 & 5.7 \\
			& HybridRAG	               & 6.3 & 8.0 & 4.4 & 10.7 & 7.2 & 7.5 & 7.0 & 7.2 \\	
			& Hybrid-RACA w/o FT	   & 4.6 & 9.0 & 4.1 & 10.9 & 5.6 & 8.9 & 5.9 & 7.1 \\
            & Hybrid-RACA FT         & 4.4 & 13.8 & 3.7 & 16.8 & 5.3 & 14.8 & 5.5 & 12.5 \\
			& Hybrid-RACA IT		   & \textbf{3.3} & \textbf{22.9} & \textbf{2.9} & \textbf{24.2} & \textbf{3.8} & \textbf{20.2} & \textbf{4.4} & \textbf{20.4} \\
			\midrule
            \multirow{4}{2.5em}{OPT-350M}

            & Vanilla OPT	        & 7.4 & 5.9 & 6.2 & 10.3 & 6.4 & 8.5 & 7.7 & 6.3 \\		
			& HybridRAG	                & 5.5 & 9.1 & 3.7 & 12.4 & 6.1 & 8.4 & 5.8 & 8.5 \\		
			& Hybrid-RACA w/o FT	    & 4.1 & 12.5 & 3.5 & 12.6 & 4.8 & 11.6 & 5.0 & 9.9 \\	
			& Hybrid-RACA FT		    & 4.2 & 13.3 & 3.5 & 17.9 & 5.1 & 13.3 & 5.2 & 13.4 \\
            & Hybrid-RACA IT          & \textbf{3.1} & \textbf{24.7} & \textbf{2.7} & \textbf{25.5} & \textbf{3.7} & \textbf{20.7} & \textbf{4.2} & \textbf{20.8} \\

			\bottomrule
		\end{tabularx}
\end{small}
\caption{Out-of-domain evaluation of Hybrid-RACA performance}
\label{table:utility_pile}
\end{table*}

\begin{table}[th!]
\centering
\begin{small}
\begin{tabular}{c|c}
\toprule
     Model &  GPT Score\\
     \midrule
     GPT3.5 & \textbf{7.73} \\
     Vanilla OPT-125M & 2.20 \\
     Vanilla OPT-350M & 2.60 \\
     \midrule
    Hybrid-RACA IT OPT-125M & 5.27 \\
     Hybrid-RACA IT OPT-350M & \textbf{5.49} \\
     \bottomrule
\end{tabular}
\caption{LLM evaluation of text completion quality}
\label{table:gpt_score}
\end{small}

\end{table}


\subsection{Experimental Setup} \label{sec:experiment_setup}

\paragraph{Datasets and Labels}
We train our models on WikiText-103 \citep{merity2016pointer} and evaluate them on the text prediction task on five datasets, including in-domain evaluation on WikiText-103, and out-of-domain evaluation on Enron Emails \cite{klimt2004enron}, HackerNews\footnote{\url{https://news.ycombinator.com/}}, NIH ExPorter\footnote{\url{https://exporter.nih.gov/}}, and Youtube Subtitles \citep{pile}, covering a diverse range of domains.
We use an LLM to generate ground truth labels for evaluation.

\paragraph{Evaluation Metrics}
To evaluate utility, we use standard metrics including perplexity (PPL) \citep{jelinek1977perplexity}, GLEU \citep{wu2016googles}, BLEU \citep{Papineni02bleu:a}, ROUGE \citep{lin-2004-rouge}, METEOR \citep{banarjee2005}, and BERTScore \citep{bert-score}. We calculate perplexity by measuring how well the model predicts the labels given the prompts.
We use other metrics to measure the degree of similarity between model predictions and the labels. In addition, we evaluated the completion quality of 100 sampled data points using GPT-4-turbo, rating on a scale of 1-10. To evaluate the inference latency of our system, we measure the average running time.

\paragraph{Implementation Details} \label{sec:implementation_details}

For the client model, we compare two small OPT models \citep{zhang2022opt}: OPT-125M and OPT-350M. Both models are decoder-only transformers that are small enough to run with limited latency budget. We employ greedy search for decoding. For the LLM, we use the GPT-3.5 text-davinci-003 model\footnote{With OpenAI API \url{https://platform.openai.com/docs/models/gpt-3.5}, $\mathrm{temperature}=0$, $\mathrm{top}\_p=1$}. We set max new tokens to 44 for both label generation and text prediction. For document retrieval, we use the Faiss library \citep{johnson2019billion} and set $k=3$ after a hyperparameter search on WikiText data.

For latency evaluation, we deploy the client models on two different machines: a GPU machine with an 11GB Nvidia Tesla K80 GPU, and a laptop without a GPU. We set max new tokens to 15 for latency evaluation.

\begin{figure*}[t]
\begin{subfigure}{.24\linewidth}
  \includegraphics[trim=1.5cm 0.1cm 0.1cm 0.1cm, width=0.9\linewidth]{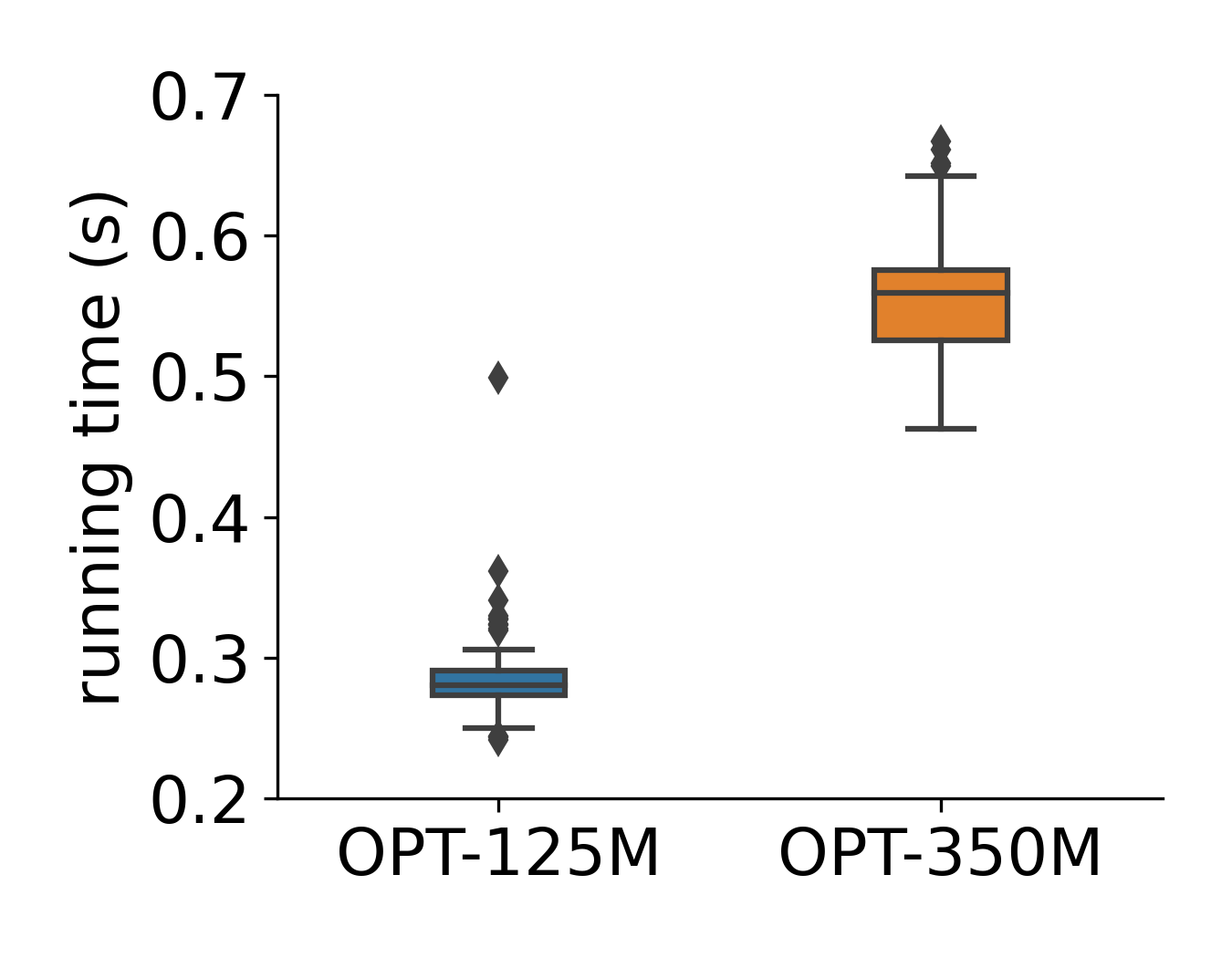}
  \caption{125M vs 350M}
  \label{fig:latency_opt125_350}
\end{subfigure}
\begin{subfigure}{.26\linewidth}
  \includegraphics[trim=1.5cm 0.1cm 0.1cm 0.1cm, width=0.9\linewidth]{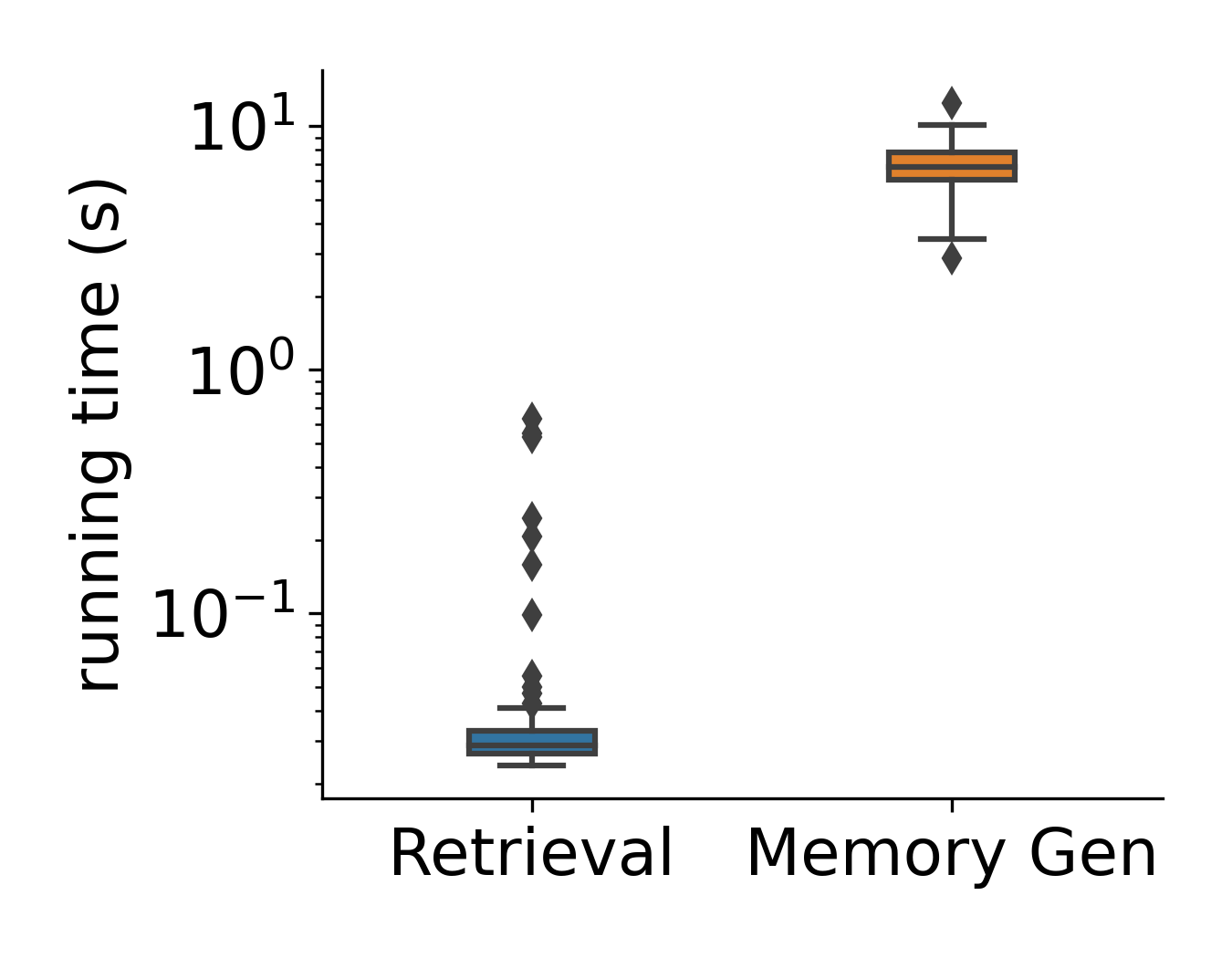}
  \caption{retrieval vs memory}
  \label{fig:latency_retr_memgen}
\end{subfigure}
\begin{subfigure}{.24\linewidth}
  \includegraphics[trim=1.8cm 0.1cm 0.1cm 0.1cm, width=0.9\linewidth]{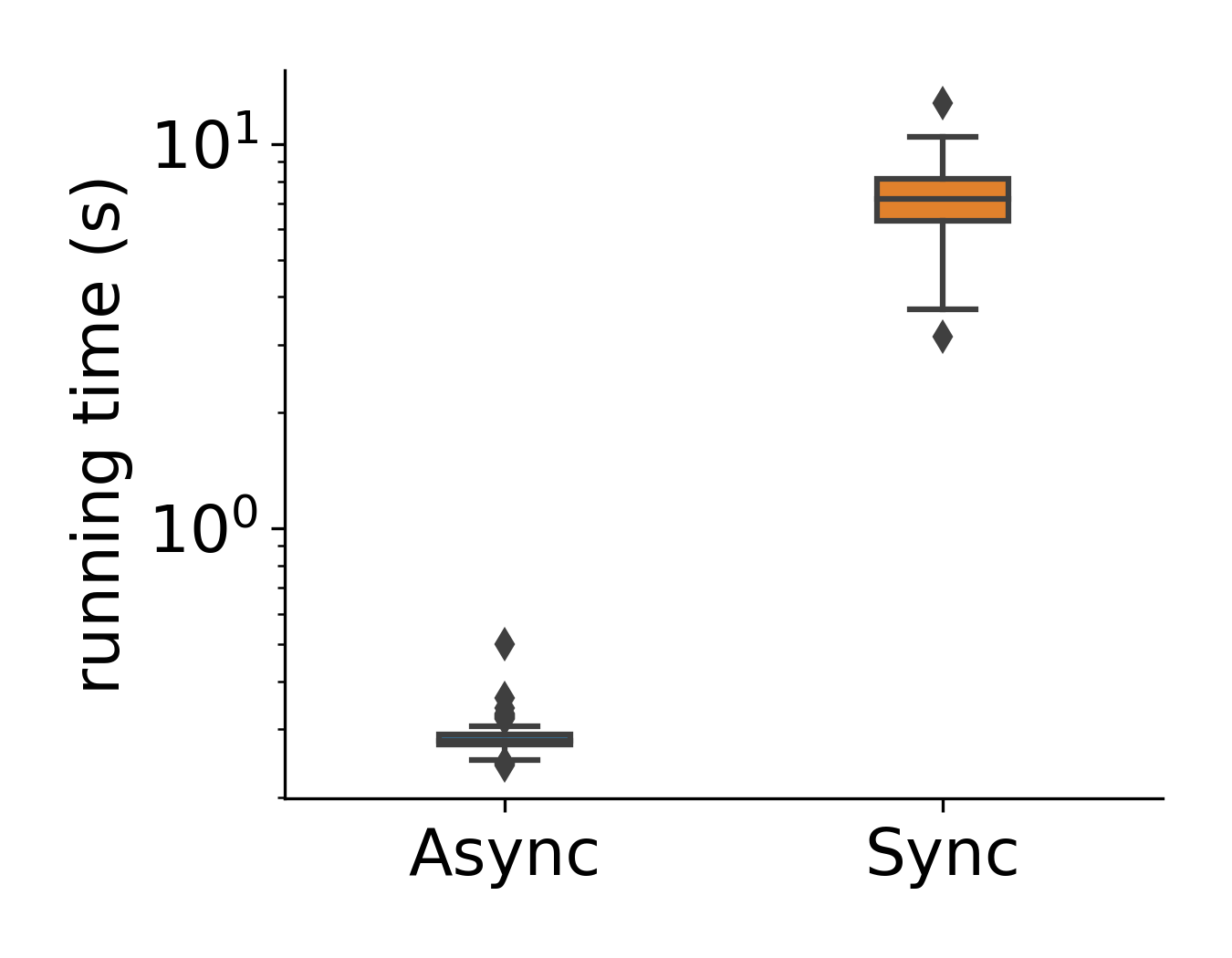}
  \caption{async vs sync}
  \label{fig:latency_sync_async}
\end{subfigure}
\begin{subfigure}{.24\linewidth}
  \includegraphics[trim=1.4cm 0.1cm 0.4cm 0.1cm, width=0.9\linewidth]{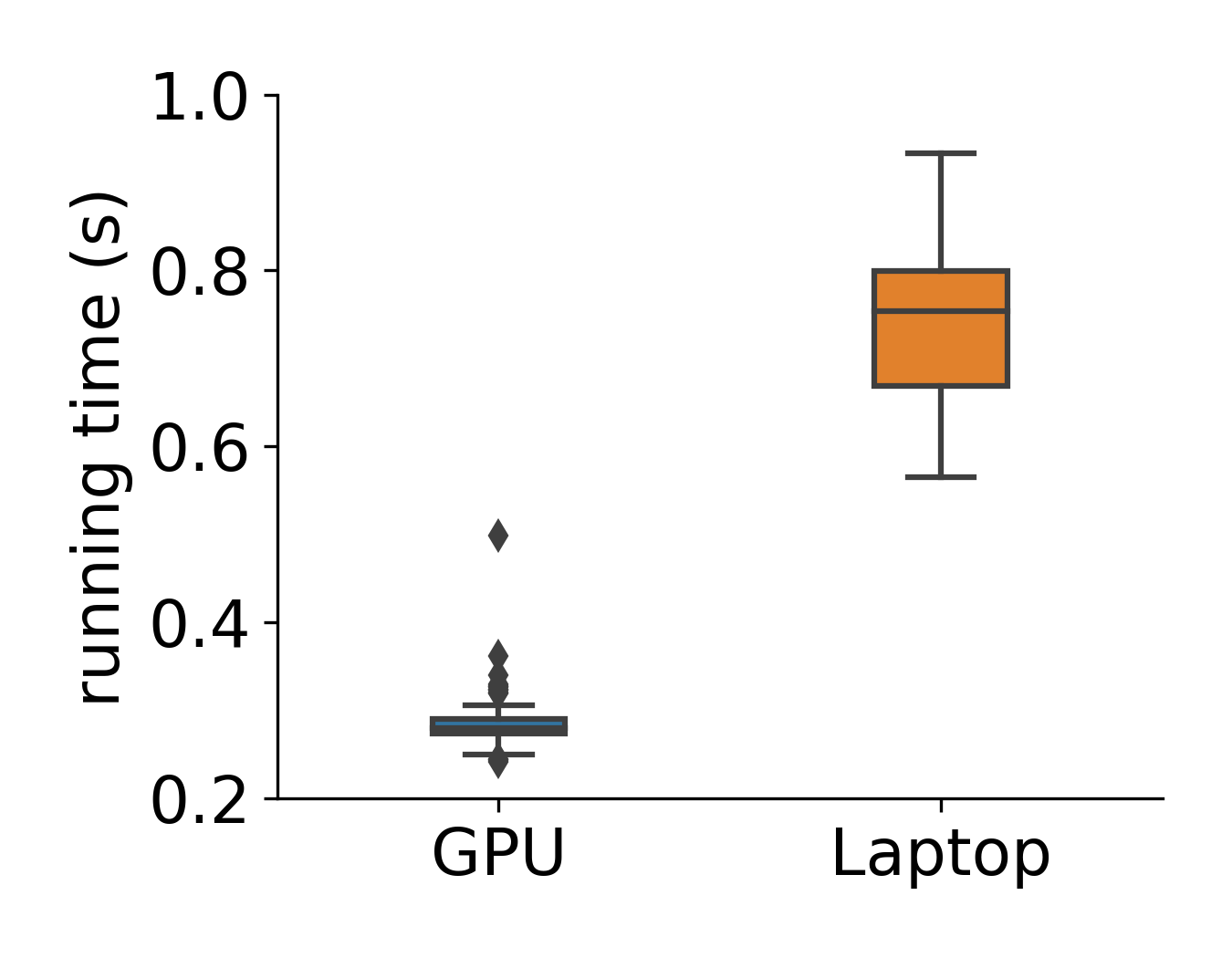}
  \caption{GPU vs laptop}
  \label{fig:latency_laptop_gpu}
\end{subfigure}

\caption{Inference latency for client inference, retrieval and memory generation on multiple devices}
\label{fig:latency}

\end{figure*}

\begin{figure*}[t]
\begin{subfigure}{.245\linewidth}
  \resizebox{\linewidth}{!}{\input{figures/async_ppl.tex}}
  \caption{PPL (async labels)}
  \label{fig:async_ppl}
\end{subfigure}
\begin{subfigure}{.245\linewidth}
  \resizebox{\linewidth}{!}{\input{figures/async_gleu.tex}}
  \caption{GLEU (async labels)}
  \label{fig:async_gleu}
\end{subfigure}
\begin{subfigure}{.245\linewidth}
  \resizebox{\linewidth}{!}{\input{figures/sync_ppl.tex}}
  \caption{PPL (sync labels)}
  \label{fig:sync_ppl}
\end{subfigure}
\begin{subfigure}{.245\linewidth}
  \resizebox{\linewidth}{!}{\input{figures/sync_gleu.tex}}
  \caption{GLEU (sync labels)}
  \label{fig:sync_gleu}
\end{subfigure}

\caption{Hybrid-RACA performance with asynchronous memory update.}
\label{fig:async}

\end{figure*}

\paragraph{Baseline Methods}
We compare our approach against four baselines. We ensure a fair comparison by regenerating labels for each baseline, based on the memory used by that baseline.

\textit{Vanilla OPT} - A vanilla client model for text prediction without additional memory from the cloud.

\textit{Hybrid-RAG} - The RAG approach \citep{lewis2020retrieval} can be turned into a hybrid setup with our system. In this setting, we retrieve and feed the full retrieved text to the client model.\footnote{This only works if the documents are sufficiently short to fit in the limited input context of the client model.}
    
\textit{Hybrid-RACA w/o FT} and \textit{Hybrid-RACA FT} - To assess the efficacy of our instruction-tuned client model, we examine two variants of the client model, one without finetuning (Hybrid-RACA w/o FT) and one finetuned to use the memory to predict the original remaining text (Hybrid-RACA FT).


\begin{figure*}[t]
    \centering
    \includegraphics[width=0.8\textwidth]{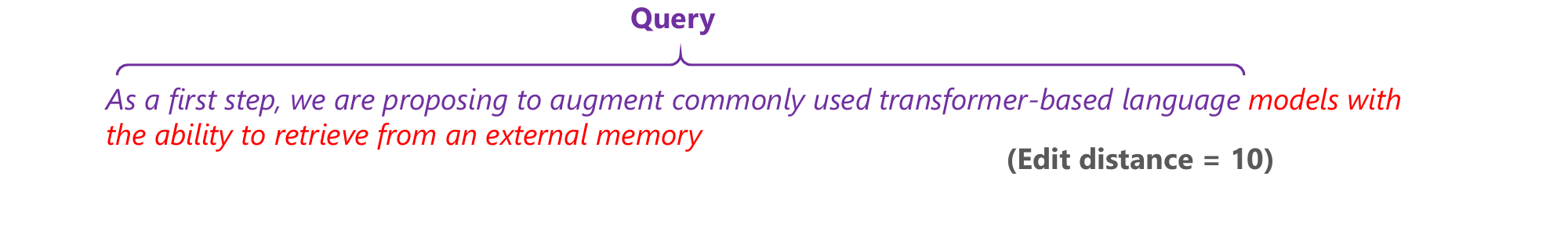}
    \caption{An example of setting edit distance threshold = 10 in asynchronous memory update. In this setting, text prediction is generated from the entire prompt, but only the beginning part is used for memory generation.}
    \label{fig:edit}
\end{figure*}
\subsection{Experimental Results} \label{sec:exp_results}

\subsubsection{Utility}
\label{sec:utility}
Table \ref{table:utility_wikitext} presents the performance of the models on WikiText-103. Table \ref{table:utility_pile} presents the perplexity and GLEU scores on the other four datasets. The results show that our approach outperforms all baselines and generalizes well to out-of-domain data. The HybridRAG approach outperforms a vanilla OPT baseline with retrieval augmentation, and the Hybrid-RACA w/o FT model improves upon it by using the LLM to extract key takeaways from retrieved data. This indicates that the representation of the context is vital to client model performance. Furthermore, our final model, \textit{Hybrid-RACA IT} (Instruction-tuned Hybrid-RACA), shows the best performance, suggesting that instruction-tuning helps the model better leverage context. Further, OPT-350M based models consistently outperform OPT-125M ones, showing that model size is critical to its overall performance. Table \ref{table:gpt_score} shows the evaluation results from GPT-4-turbo, demonstrating that Hybrid-RACA significantly enhances text completion quality.

\subsubsection{Inference Latency}
\label{sec:latency}
We performed a latency evaluation for Hybrid-RACA. Fig.\ref{fig:latency_opt125_350} shows the run times for the client models on a GPU machine. Unsurprisingly, OPT-125M is 49.3\% faster compared to OPT-350M.
Fig.\ref{fig:latency_retr_memgen} presents the run times for retrieval and memory generation steps, showing that memory generation with LLM consumes the majority time for memory preparation.
Fig.\ref{fig:latency_sync_async} compares our asynchronous Hybrid-RACA (OPT-125M) to a synchronous approach by directly calling GPT-3.5 and a retriever for composition assistance. Notably, our approach showcases an impressive speed improvement, achieving a remarkable 138x faster performance compared to the synchronous approach. 
Fig.\ref{fig:latency_laptop_gpu} compares the run times of Hybrid-RACA OPT-125M on a GPU machine and a laptop without GPU. It shows that our approach can be deployed on edge devices without GPUs, although slower. 

Notably, we didn't use caching or quantization. These methods are orthogonal to our work and can be used in conjunction to further improve the speed.

\subsubsection{Asynchronous Memory Update}
\label{sec:async}
Fig.\ref{fig:async} illustrates the impact of asynchronous memory update on model utility. To measure this effect, we conducted an experiment in which we gradually increased the edit distance threshold that determines how often the client model requests for memory updates. For each prompt, we use the beginning part of the prompt as the query for memory generation and the entire prompt for text prediction, mimicking the case where the memory lags behind the current input context due to asynchronous communication between client and cloud. Figure \ref{fig:edit} demonstrates how we set the edit distance threshold in async memory update. 

Fig.\ref{fig:async_ppl} and Fig.\ref{fig:async_gleu} show the trend in perplexity and GLEU scores with increased edit distance threshold, evaluated against GPT3.5 generated labels with the same asynchronous setup. Model utility remains relatively stable in perplexity with a deceasing trend in GLEU compared to LLMs. Fig.\ref{fig:sync_ppl} and Fig.\ref{fig:sync_gleu} show the scores of the client model under the asynchronous setup, evaluated against labels generated in an ideal synchronous memory update setup, where the memory is created using the entire prompt without lag. 
Due to the difference in the freshness of the memory, there is a larger gap between the asynchronous predictions and the synchronous labels. 
As the edit distance threshold increases, the memory becomes less up-to-date, resulting in a decline in model utility. Nevertheless, it still significantly outperformed the baselines.

%% file: figures/async_ppl.tex
\begin{tikzpicture}

\definecolor{darkgray176}{RGB}{176,176,176}
\definecolor{darkorange25512714}{RGB}{255,127,14}
\definecolor{lightgray204}{RGB}{204,204,204}
\definecolor{steelblue31119180}{RGB}{31,119,180}

\begin{axis}[
legend cell align={left},
legend style={fill opacity=0.8, draw opacity=1, text opacity=1, draw=lightgray204},
tick align=outside,
tick pos=left,
x grid style={darkgray176},
xlabel={Edit distance},
xmin=4.25, xmax=20.75,
xtick style={color=black},
y grid style={darkgray176},
ylabel={Perplexity},
ymin=2, ymax=3,
ytick style={color=black}
]
\addplot [semithick, steelblue31119180, mark=*, mark size=3, mark options={solid}]
table {%
5 2.51705484121686
10 2.50946946794304
15 2.52859648335547
20 2.54251015024738
};
\addlegendentry{opt-125m}
\addplot [semithick, darkorange25512714, mark=x, mark size=3, mark options={solid}]
table {%
5 2.33769063107641
10 2.32922815150913
15 2.34891448910029
20 2.36339371375384
};
\addlegendentry{opt-350m}
\end{axis}

\end{tikzpicture}

%% file: figures/async_gleu.tex
\begin{tikzpicture}

\definecolor{darkgray176}{RGB}{176,176,176}
\definecolor{darkorange25512714}{RGB}{255,127,14}
\definecolor{lightgray204}{RGB}{204,204,204}
\definecolor{steelblue31119180}{RGB}{31,119,180}

\begin{axis}[
legend cell align={left},
legend style={fill opacity=0.8, draw opacity=1, text opacity=1, draw=lightgray204},
tick align=outside,
tick pos=left,
x grid style={darkgray176},
xlabel={Edit distance},
xmin=4.25, xmax=20.75,
xtick style={color=black},
y grid style={darkgray176},
ylabel={GLEU},
ymin=28, ymax=34,
ytick style={color=black}
]
\addplot [semithick, steelblue31119180, mark=*, mark size=3, mark options={solid}]
table {%
5 30.6665905876413
10 29.963496389772
15 29.3574295748123
20 28.9408951666501
};
\addlegendentry{opt-125m}
\addplot [semithick, darkorange25512714, mark=x, mark size=3, mark options={solid}]
table {%
5 32.9586762540068
10 32.3506455779051
15 32.0611428139934
20 31.5524332521483
};
\addlegendentry{opt-350m}
\end{axis}

\end{tikzpicture}

%% file: figures/sync_ppl.tex
\begin{tikzpicture}

\definecolor{darkgray176}{RGB}{176,176,176}
\definecolor{darkorange25512714}{RGB}{255,127,14}
\definecolor{lightgray204}{RGB}{204,204,204}
\definecolor{steelblue31119180}{RGB}{31,119,180}

\begin{axis}[
legend cell align={left},
legend style={fill opacity=0.8, draw opacity=1, text opacity=1, draw=lightgray204},
tick align=outside,
tick pos=left,
x grid style={darkgray176},
xlabel={Edit distance},
xmin=4.25, xmax=20.75,
xtick style={color=black},
y grid style={darkgray176},
ylabel={Perplexity},
ymin=5.5, ymax=7.5,
ytick style={color=black}
]
\addplot [semithick, steelblue31119180, mark=*, mark size=3, mark options={solid}]
table {%
5 6.2
10 6.22
15 6.38
20 6.54
};
\addlegendentry{opt-125m}
\addplot [semithick, darkorange25512714, mark=x, mark size=3, mark options={solid}]
table {%
5 6.28
10 6.31
15 6.55
20 6.73
};
\addlegendentry{opt-350m}
\end{axis}

\end{tikzpicture}

%% file: figures/sync_gleu.tex
\begin{tikzpicture}

\definecolor{darkgray176}{RGB}{176,176,176}
\definecolor{darkorange25512714}{RGB}{255,127,14}
\definecolor{lightgray204}{RGB}{204,204,204}
\definecolor{steelblue31119180}{RGB}{31,119,180}

\begin{axis}[
legend cell align={left},
legend style={fill opacity=0.8, draw opacity=1, text opacity=1, draw=lightgray204},
tick align=outside,
tick pos=left,
x grid style={darkgray176},
xlabel={Edit distance},
xmin=4.25, xmax=20.75,
xtick style={color=black},
y grid style={darkgray176},
ylabel={GLEU},
ymin=18, ymax=24,
ytick style={color=black}
]
\addplot [semithick, steelblue31119180, mark=*, mark size=3, mark options={solid}]
table {%
5 21.5
10 20.56
15 19.82
20 18.99
};
\addlegendentry{opt-125m}
\addplot [semithick, darkorange25512714, mark=x, mark size=3, mark options={solid}]
table {%
5 22.42
10 21.58
15 21.15
20 19.98
};
\addlegendentry{opt-350m}
\end{axis}

\end{tikzpicture}

%% file: sections/5_conclusion.tex
\section{Conclusion}

In this paper, we propose Hybrid-RACA, a novel hybrid retrieval-augmented generation system for real-time composition assistance. By integrating LLM-enhanced memory into our instruction-tuned client model with asynchronous update, we show with experiment results on multiple datasets that our hybrid approach enables substantial utility improvements over smaller language models while maintaining inference efficiency, making it a valuable solution for real-time tasks.


%% file: sections/6_appendix.tex
\section{More results on utility evaluation}
\label{sec:pile}
The results of the model utility on Enron Emails, NIH ExPorter, HackerNews, and YouTubeSubtitles datasets evaluated in all seven metrics are presented in Tables \ref{table:utility_pile_350M} and \ref{table:utility_pile_125M}. We can observe that our model consistently outperforms all the other baselines.
\begin{table*}[ht!]
\centering
\begin{small}
	\small
    \scalebox{0.9}{
    \begin{tabularx}{1.1\textwidth}{c|c|c *{6}{Y}}
    \toprule
     &   &  PPL  &  GLEU  &  BLEU-4  &  ROUGE-1  &  ROUGE-L  &  METEOR  &  BERTScore  \\
    \midrule
    \multirow{5}{4em}{Enron Emails}
     &  Vanilla OPT	       & 7.4 & 5.9 & 2.7 & 17.3 & 14.3 & 13.2 & 80.2 \\
     &  HybridRAG	               & 5.5 & 9.1 & 6.6 & 21.7 & 18.1 & 17.0 & 80.6 \\
     &  Hybrid-RACA w/o FT	       & 4.1 & 12.5 & 10.8 & 25.3 & 21.6 & 21.1 & 81.8 \\
     &  Hybrid-RACA FT			   & 4.2 & 13.3 & 11.6 & 26.5 & 22.1 & 22.8 & 83.1 \\
     &  Hybrid-RACA IT			   & \textbf{3.1} & \textbf{24.7} & \textbf{22.7} & \textbf{43.9} & \textbf{35.4} & \textbf{39.6} & \textbf{87.9} \\

    \midrule
    \multirow{5}{4em}{NIH ExPorter}
     &  Vanilla OPT	       & 6.2 & 10.3 & 5.4 & 27.7 & 22.3 & 19.6 & 85.3 \\
     &  HybridRAG	               & 3.7 & 12.4 & 8.9 & 30.2 & 24.5 & 23.3 & 85.8 \\
     &  Hybrid-RACA w/o FT	       & 3.5 & 12.6 & 9.3 & 30.0 & 24.6 & 23.7 & 85.7 \\
     &  Hybrid-RACA FT			   & 3.5 & 17.9 & 15.4 & 36.5 & 29.4 & 30.6 & 87.2 \\
     &  Hybrid-RACA IT			   & \textbf{2.7} & \textbf{25.5} & \textbf{23.2} & \textbf{45.9} & \textbf{37.2} & \textbf{41.2} & \textbf{89.2} \\

    \midrule
    \multirow{5}{4em}{Hacker News}
     &  Vanilla OPT	       & 6.4 & 8.5 & 5.0 & 24.7 & 20.5 & 16.3 & 84.9 \\
     &  HybridRAG	               & 6.1 & 8.4 & 5.6 & 22.4 & 18.9 & 14.7 & 83.9 \\
     &  Hybrid-RACA w/o FT	       & 4.8 & 11.6 & 9.2 & 27.0 & 22.6 & 19.4 & 84.9 \\
     &  Hybrid-RACA FT			   & 5.1 & 13.3 & 11.4 & 28.2 & 23.0 & 21.6 & 84.8 \\
     &  Hybrid-RACA IT			   & \textbf{3.7} & \textbf{20.7} & \textbf{18.2} & \textbf{40.3} & \textbf{31.6} & \textbf{35.3} & \textbf{87.8} \\

    \midrule
    \multirow{5}{4em}{Youtube Subtitles}
     &  Vanilla OPT	       & 7.7 & 6.3 & 2.7 & 17.8 & 15.1 & 13.8 & 82.2 \\
     &  HybridRAG	               & 5.8 & 8.5 & 5.2 & 22.3 & 18.1 & 17.4 & 83.5 \\
     &  Hybrid-RACA w/o FT	       & 5.0 & 9.9 & 7.4 & 22.1 & 18.4 & 18.1 & 83.2 \\
     &  Hybrid-RACA FT			   & 5.2 & 13.4 & 11.0 & 27.1 & 22.0 & 23.0 & 84.5 \\
     &  Hybrid-RACA IT			   & \textbf{4.2} & \textbf{20.8} & \textbf{18.3} & \textbf{39.2} & \textbf{30.7} & \textbf{34.7} & \textbf{87.4} \\

    \bottomrule
    \end{tabularx}}

    \caption{Comparison of the utility performance of the OPT-350M-based Hybrid-RACA models and baselines on four datasets}
    \label{table:utility_pile_350M}
\end{small}
\end{table*}

\begin{table*}[ht!]
\centering
\begin{small}
	\small
    \scalebox{0.9}{
    \begin{tabularx}{1.1\textwidth}{c|c|c *{6}{Y}}
    \toprule
     &   &  PPL  &  GLEU  &  BLEU-4  &  ROUGE-1  &  ROUGE-L  &  METEOR  &  BERTScore  \\
    \midrule
    \multirow{5}{4em}{Enron Emails}
     &  Vanilla OPT	       & 8.5 & 5.8 & 2.6 & 17.4 & 14.7 & 13.5 & 80.1 \\
     &  Hybrid-RACA	               & 6.3 & 8.0 & 5.9 & 20.0 & 17.1 & 15.4 & 79.6 \\
     &  Hybrid-RACA w/o FT	       & 4.6 & 9.0 & 6.9 & 20.8 & 17.9 & 16.9 & 80.9 \\
     &  Hybrid-RACA FT			   & 4.4 & 13.8 & 12.1 & 26.9 & 22.6 & 23.3 & 83.3 \\
     &  Hybrid-RACA IT			   & \textbf{3.3} & \textbf{22.9} & \textbf{20.9} & \textbf{41.6} & \textbf{33.3} & \textbf{37.1} & \textbf{86.9} \\
     
    \midrule
    \multirow{5}{4em}{NIH ExPorter}
     &  Vanilla OPT	       & 7.4 & 9.3 & 4.5 & 25.9 & 21.1 & 18.3 & 84.8 \\
     &  HybridRAG	               & 4.4 & 10.7 & 7.1 & 27.4 & 22.5 & 20.8 & 84.9 \\
     &  Hybrid-RACA w/o FT	       & 4.1 & 10.9 & 7.7 & 26.9 & 22.5 & 21.0 & 84.9 \\
     &  Hybrid-RACA FT			   & 3.7 & 16.8 & 14.4 & 34.9 & 28.3 & 29.3 & 86.7 \\
    &  Hybrid-RACA IT			   & \textbf{2.9} & \textbf{24.2} & \textbf{21.9} & \textbf{44.3} & \textbf{35.6} & \textbf{39.4} & \textbf{88.8} \\

    \midrule
    \multirow{5}{4em}{Hacker News}
     &  Vanilla OPT	       & 7.5 & 8.0 & 4.6 & 23.1 & 19.4 & 15.3 & 84.1 \\
     &  Hybrid-RACA	               & 7.2 & 7.5 & 4.8 & 20.9 & 18.0 & 13.4 & 83.4 \\
     &  Hybrid-RACA w/o FT	      & 5.6 & 8.9 & 6.4 & 22.6 & 19.4 & 15.3 & 83.8 \\
     &  Hybrid-RACA FT			   & 5.3 & 14.8 & 12.8 & 30.3 & 24.8 & 23.6 & 85.4 \\
        &  Hybrid-RACA IT			   & \textbf{3.8} & \textbf{20.2} & \textbf{18.0} & \textbf{39.3} & \textbf{30.8} & \textbf{33.3} & \textbf{87.5} \\

    \midrule
    \multirow{5}{4em}{Youtube Subtitles}
     &  Vanilla OPT	       & 9.2 & 5.7 & 2.2 & 16.7 & 14.2 & 13.1 & 82.6 \\
     &  Hybrid-RACA	               & 7.0 & 7.2 & 4.1 & 19.4 & 16.7 & 15.5 & 82.9 \\
     &  Hybrid-RACA w/o FT	      & 5.9 & 7.1 & 4.0 & 18.2 & 15.7 & 14.8 & 82.1 \\
     &  Hybrid-RACA FT			   & 5.5 & 12.5 & 9.9 & 26.1 & 21.4 & 22.5 & 84.6 \\
     &  Hybrid-RACA IT			   & \textbf{4.4} & \textbf{20.4} & \textbf{17.8} & \textbf{38.7} & \textbf{30.5} & \textbf{34.8} & \textbf{87.3} \\

    \bottomrule
    \end{tabularx}}

    \caption{Comparison of the utility performance of the OPT-125M-based Hybrid-RACA models and baselines on four datasets}
    \label{table:utility_pile_125M}
\end{small}
\end{table*}


\section{Template used to calculate GPT-score}

We use the following template to instruct GPT-4-turbo to evaluate the performance of of the models:

\texttt{
Please act as an impartial judge and evaluate the quality of the text completion provided by an AI assistant to the text prompt displayed below. For this evaluation, you should primarily consider the following criteria:
\newline
relevance: Is the completion relevant to the prompt? Is the completion a fluent continuation from the prompt?
\newline
correctness: Is the completion correct and factual?
\newline
fluency: Is the completion fluent, free of grammatical errors, and devoid of redundant repetitions? Please note that it is acceptable for the completion to stop abruptly before the end of a sentence.
\newline
Begin your evaluation by providing a short explanation. Be as objective as possible. After providing your explanation, you must rate the response on a scale of 1 to 10 by strictly following this format: "[[rating]]", for example: "Rating: [[5]]"
\newline
\newline
[Text Prompt]
\newline
\{prompt\}
\newline
\newline
[The Start of Assistant's Completion]
\{completion\}
\newline
[The End of Assistant's Completion]
}

\section{Examples of the model completions} 
\label{sec:case_study}

Table \ref{tab:completions} shows a working example for Hybrid-RACA models and Table \ref{tab:completions_2} and \ref{tab:completions_3} show examples of failing cases.

Table \ref{tab:completions_2} is a failing case for both OPT-125M and OPT-350M Hybrid-RACA models. In this case, the memory doesn't contain the information needed to complete the text. As a large language model, GPT3.5 is capable of ignoring the memory and using its parametric memory to generate the completion. However, the smaller client models tend to pick entities present in the memory for text generation despite that the resulting completion is not factually accurate. Table \ref{tab:completions_3} shows an example of working case for Hybrid-RACA OPT-350M IT model, but a failing case for the OPT-125M based model. In this case, the memories are bullet points generated from several document chunks; the client model with limited reasoning abilities does not allow them to process the memories extensively and reorganize them. We've noticed that when the small OPT client models combine information from different bits of the memories, they usually end up fabricating wrong facts.

Improving the memory generator by reducing duplicate information, and enhancing the reasoning abilities of the client model or encouraging it to select more useful memories and ignore irrelevant content would be some of the ways to address these failing cases and limitations.

\begin{table*}[t]
\centering
\small
\begin{tabular}{p{0.25\textwidth} p{0.7\textwidth}}
\toprule
prompt & Air Mata Iboe (Perfected spelling: Air \\ \midrule
GPT-3.5 generated reference & \textcolor{green}{ Mata Ibu) was a musical extravaganza released in December 1941. It was distributed by Columbia Pictures and premiered at Sampoerna Theatre in Surabaya. It was advertised under the Dutch title Tran} \\ \midrule
Hybrid-RACA OPT-125M IT & \textcolor{green}{ Mata Iboe) was released in December 1941 and was distributed by Columbia Pictures. It was advertised under the Dutch title Tranen Eener Moeder and was promoted as a "musical extravaganza".} \\ \midrule
Hybrid-RACA OPT-350M IT & \textcolor{green}{ Mata Iboe) was released in December 1941 and was distributed by Columbia Pictures. It was advertised under the Dutch title Tranen Eener Moeder and was promoted as a "musical extravaganza".} \\ \midrule
\bottomrule
\end{tabular}
\caption{A working case for Hybrid-RACA models}
\label{tab:completions}
\end{table*}

\begin{table*}[t]
\centering
\small
\begin{tabular}{p{0.25\textwidth} p{0.7\textwidth}}
\toprule
prompt & Viva el Príncipe (English: Long Live the Prince) is the thirteenth studio album by Mexican recording artist \\ \midrule
GPT-3.5 generated reference & \textcolor{green}{ Cristian Castro. It was released on November 30, 2010 and was met with critical acclaim. The album peaked at No. 5 on the Argentine Chamber of Phonograms and Videograms Producers pop charts and received gold} \\ \midrule
Hybrid-RACA OPT-125M IT & \textcolor{red}{ David Jeffries. It was released on November 30, 2010 worldwide and peaked at No. 5 on the Argentine Chamber of Phonograms and Videograms Producers pop charts. The album was certified diamond by AMPROF} \\ \midrule
Hybrid-RACA OPT-350M IT & \textcolor{red}{ David Jeffries. It was released on November 30, 2010 worldwide and peaked at No. 5 on the Argentine Chamber of Phonograms and Videograms Producers pop charts. It was certified diamond by AMPROFON} \\ \midrule
\bottomrule
\end{tabular}
\caption{A failing case for Hybrid-RACA models.}
\label{tab:completions_2}
\end{table*}

\begin{table*}[t]
\centering
\small
\begin{tabular}{p{0.25\textwidth} p{0.7\textwidth}}
\toprule
prompt & "Make Me Like You" is a song by American singer Gwen Stefani \\ \midrule
GPT-3.5 generated reference & \textcolor{green}{, released as the second single from her third studio album This Is What the Truth Feels Like. The track was written by Stefani, Justin Tranter, Julia Michaels, Mattias Larsson and Robin Fredriks} \\ \midrule
Hybrid-RACA OPT-125M IT & \textcolor{red}{ that has become a classic. Released digitally on February 12, 2016 as the album's second single, it was serviced to mainstream radio on February 16, 2016 in the United States. The track was written by Stefani} \\ \midrule
Hybrid-RACA OPT-350M IT & \textcolor{green}{ that was released digitally on February 12, 2016. It was written by Stefani, Justin Tranter, Julia Michaels, Mattias Larsson and Robin Fredriksson under their stage name Mattman \& Robin. The} \\ \midrule
\bottomrule
\end{tabular}
\caption{A working case for Hybrid-RACA OPT-350M IT but failed for other variants.}
\label{tab:completions_3}
\end{table*}

%% file: main.bbl
\begin{thebibliography}{30}
\expandafter\ifx\csname natexlab\endcsname\relax\def\natexlab#1{#1}\fi

\bibitem[{Banerjee and Lavie(2005)}]{banarjee2005}
Satanjeev Banerjee and Alon Lavie. 2005.
\newblock \href {https://www.aclweb.org/anthology/W05-0909} {{METEOR}: An automatic metric for {MT} evaluation with improved correlation with human judgments}.
\newblock In \emph{Proceedings of the {ACL} Workshop on Intrinsic and Extrinsic Evaluation Measures for Machine Translation and/or Summarization}, pages 65--72, Ann Arbor, Michigan. Association for Computational Linguistics.

\bibitem[{Bonawitz et~al.(2019)Bonawitz, Eichner, Grieskamp, Huba, Ingerman, Ivanov, Kiddon, Kone{\v{c}}n{\`y}, Mazzocchi, McMahan et~al.}]{bonawitz2019towards}
Keith Bonawitz, Hubert Eichner, Wolfgang Grieskamp, Dzmitry Huba, Alex Ingerman, Vladimir Ivanov, Chloe Kiddon, Jakub Kone{\v{c}}n{\`y}, Stefano Mazzocchi, Brendan McMahan, et~al. 2019.
\newblock Towards federated learning at scale: System design.
\newblock \emph{Proceedings of machine learning and systems}, 1:374--388.

\bibitem[{Borgeaud et~al.(2021)Borgeaud, Mensch, Hoffmann, Cai, Rutherford, Millican, van~den Driessche, Lespiau, Damoc, Clark, de~Las~Casas, Guy, Menick, Ring, Hennigan, Huang, Maggiore, Jones, Cassirer, Brock, Paganini, Irving, Vinyals, Osindero, Simonyan, Rae, Elsen, and Sifre}]{borgeaud2021retro}
Sebastian Borgeaud, Arthur Mensch, Jordan Hoffmann, Trevor Cai, Eliza Rutherford, Katie Millican, George van~den Driessche, Jean{-}Baptiste Lespiau, Bogdan Damoc, Aidan Clark, Diego de~Las~Casas, Aurelia Guy, Jacob Menick, Roman Ring, Tom Hennigan, Saffron Huang, Loren Maggiore, Chris Jones, Albin Cassirer, Andy Brock, Michela Paganini, Geoffrey Irving, Oriol Vinyals, Simon Osindero, Karen Simonyan, Jack~W. Rae, Erich Elsen, and Laurent Sifre. 2021.
\newblock \href {http://arxiv.org/abs/2112.04426} {Improving language models by retrieving from trillions of tokens}.
\newblock \emph{CoRR}, abs/2112.04426.

\bibitem[{Fan et~al.(2021)Fan, Gardent, Braud, and Bordes}]{fan-etal-2021-augmenting}
Angela Fan, Claire Gardent, Chlo{\'e} Braud, and Antoine Bordes. 2021.
\newblock \href {https://doi.org/10.1162/tacl_a_00356} {Augmenting transformers with {KNN}-based composite memory for dialog}.
\newblock \emph{Transactions of the Association for Computational Linguistics}, 9:82--99.

\bibitem[{Gao et~al.(2020)Gao, Biderman, Black, Golding, Hoppe, Foster, Phang, He, Thite, Nabeshima, Presser, and Leahy}]{pile}
Leo Gao, Stella Biderman, Sid Black, Laurence Golding, Travis Hoppe, Charles Foster, Jason Phang, Horace He, Anish Thite, Noa Nabeshima, Shawn Presser, and Connor Leahy. 2020.
\newblock The {P}ile: An 800gb dataset of diverse text for language modeling.
\newblock \emph{arXiv preprint arXiv:2101.00027}.

\bibitem[{Guu et~al.(2020)Guu, Lee, Tung, Pasupat, and Chang}]{guu2020realm}
Kelvin Guu, Kenton Lee, Zora Tung, Panupong Pasupat, and Ming-Wei Chang. 2020.
\newblock Realm: Retrieval-augmented language model pre-training.
\newblock In \emph{Proceedings of the 37th International Conference on Machine Learning}, ICML'20. JMLR.org.

\bibitem[{Hoefler et~al.(2021)Hoefler, Alistarh, Ben-Nun, Dryden, and Peste}]{hoefler2021sparsity}
Torsten Hoefler, Dan Alistarh, Tal Ben-Nun, Nikoli Dryden, and Alexandra Peste. 2021.
\newblock Sparsity in deep learning: Pruning and growth for efficient inference and training in neural networks.
\newblock \emph{J. Mach. Learn. Res.}, 22(1).

\bibitem[{Izacard and Grave(2021)}]{izacard-grave-2021-leveraging}
Gautier Izacard and Edouard Grave. 2021.
\newblock \href {https://doi.org/10.18653/v1/2021.eacl-main.74} {Leveraging passage retrieval with generative models for open domain question answering}.
\newblock In \emph{Proceedings of the 16th Conference of the European Chapter of the Association for Computational Linguistics: Main Volume}, pages 874--880, Online. Association for Computational Linguistics.

\bibitem[{Jelinek et~al.(1977)Jelinek, Mercer, Bahl, and Baker}]{jelinek1977perplexity}
Fred Jelinek, Robert~L Mercer, Lalit~R Bahl, and James~K Baker. 1977.
\newblock Perplexity—a measure of the difficulty of speech recognition tasks.
\newblock \emph{The Journal of the Acoustical Society of America}, 62(S1):S63--S63.

\bibitem[{Johnson et~al.(2019)Johnson, Douze, and J{\'e}gou}]{johnson2019billion}
Jeff Johnson, Matthijs Douze, and Herv{\'e} J{\'e}gou. 2019.
\newblock Billion-scale similarity search with gpus.
\newblock \emph{IEEE Transactions on Big Data}, 7(3):535--547.

\bibitem[{Karpukhin et~al.(2020)Karpukhin, O{\u{g}}uz, Min, Lewis, Wu, Edunov, Chen, and Yih}]{karpukhin2020dense}
Vladimir Karpukhin, Barlas O{\u{g}}uz, Sewon Min, Patrick Lewis, Ledell Wu, Sergey Edunov, Danqi Chen, and Wen-tau Yih. 2020.
\newblock Dense passage retrieval for open-domain question answering.
\newblock \emph{arXiv preprint arXiv:2004.04906}.

\bibitem[{Khandelwal et~al.(2020)Khandelwal, Levy, Jurafsky, Zettlemoyer, and Lewis}]{Khandelwal2020Generalization}
Urvashi Khandelwal, Omer Levy, Dan Jurafsky, Luke Zettlemoyer, and Mike Lewis. 2020.
\newblock \href {https://openreview.net/forum?id=HklBjCEKvH} {Generalization through memorization: Nearest neighbor language models}.
\newblock In \emph{International Conference on Learning Representations}.

\bibitem[{Klimt and Yang(2004)}]{klimt2004enron}
Bryan Klimt and Yiming Yang. 2004.
\newblock The enron corpus: A new dataset for email classification research.
\newblock In \emph{European conference on machine learning}, pages 217--226. Springer.

\bibitem[{Kudugunta et~al.(2021)Kudugunta, Huang, Bapna, Krikun, Lepikhin, Luong, and Firat}]{kudugunta-etal-2021-beyond-distillation}
Sneha Kudugunta, Yanping Huang, Ankur Bapna, Maxim Krikun, Dmitry Lepikhin, Minh-Thang Luong, and Orhan Firat. 2021.
\newblock \href {https://doi.org/10.18653/v1/2021.findings-emnlp.304} {Beyond distillation: Task-level mixture-of-experts for efficient inference}.
\newblock In \emph{Findings of the Association for Computational Linguistics: EMNLP 2021}, pages 3577--3599, Punta Cana, Dominican Republic. Association for Computational Linguistics.

\bibitem[{Lewis et~al.(2020)Lewis, Perez, Piktus, Petroni, Karpukhin, Goyal, K{\"u}ttler, Lewis, Yih, Rockt{\"a}schel et~al.}]{lewis2020retrieval}
Patrick Lewis, Ethan Perez, Aleksandra Piktus, Fabio Petroni, Vladimir Karpukhin, Naman Goyal, Heinrich K{\"u}ttler, Mike Lewis, Wen-tau Yih, Tim Rockt{\"a}schel, et~al. 2020.
\newblock Retrieval-augmented generation for knowledge-intensive nlp tasks.
\newblock \emph{Advances in Neural Information Processing Systems}, 33:9459--9474.

\bibitem[{Lin(2004)}]{lin-2004-rouge}
Chin-Yew Lin. 2004.
\newblock \href {https://www.aclweb.org/anthology/W04-1013} {{ROUGE}: A package for automatic evaluation of summaries}.
\newblock In \emph{Text Summarization Branches Out}, pages 74--81, Barcelona, Spain. Association for Computational Linguistics.

\bibitem[{Liu et~al.(2022)Liu, Zheng, Gupta, Gaonkar, Gao, Vosoughi, Shokouhi, and Awadallah}]{liu2022knowledge}
Ruibo Liu, Guoqing Zheng, Shashank Gupta, Radhika Gaonkar, Chongyang Gao, Soroush Vosoughi, Milad Shokouhi, and Ahmed~Hassan Awadallah. 2022.
\newblock Knowledge infused decoding.
\newblock \emph{arXiv preprint arXiv:2204.03084}.

\bibitem[{Loghin et~al.(2019)Loghin, Ramapantulu, and Teo}]{loghin2019towards}
Dumitrel Loghin, Lavanya Ramapantulu, and Yong~Meng Teo. 2019.
\newblock \href {https://doi.org/10.1109/EDGE.2019.00029} {Towards analyzing the performance of hybrid edge-cloud processing}.
\newblock In \emph{2019 IEEE International Conference on Edge Computing (EDGE)}, pages 87--94.

\bibitem[{Matsubara et~al.(2022)Matsubara, Levorato, and Restuccia}]{matsubara2022split}
Yoshitomo Matsubara, Marco Levorato, and Francesco Restuccia. 2022.
\newblock \href {https://doi.org/10.1145/3527155} {Split computing and early exiting for deep learning applications: Survey and research challenges}.
\newblock \emph{ACM Comput. Surv.}, 55(5).

\bibitem[{Merity et~al.(2016)Merity, Xiong, Bradbury, and Socher}]{merity2016pointer}
Stephen Merity, Caiming Xiong, James Bradbury, and Richard Socher. 2016.
\newblock \href {http://arxiv.org/abs/1609.07843} {Pointer sentinel mixture models}.

\bibitem[{Osia et~al.(2020)Osia, Shahin~Shamsabadi, Sajadmanesh, Taheri, Katevas, Rabiee, Lane, and Haddadi}]{osia2020ahybrid}
Seyed~Ali Osia, Ali Shahin~Shamsabadi, Sina Sajadmanesh, Ali Taheri, Kleomenis Katevas, Hamid~R. Rabiee, Nicholas~D. Lane, and Hamed Haddadi. 2020.
\newblock \href {https://doi.org/10.1109/JIOT.2020.2967734} {A hybrid deep learning architecture for privacy-preserving mobile analytics}.
\newblock \emph{IEEE Internet of Things Journal}, 7(5):4505--4518.

\bibitem[{Papineni et~al.(2002)Papineni, Roukos, Ward, and jing Zhu}]{Papineni02bleu:a}
Kishore Papineni, Salim Roukos, Todd Ward, and Wei jing Zhu. 2002.
\newblock Bleu: a method for automatic evaluation of machine translation.
\newblock pages 311--318.

\bibitem[{Shi et~al.(2023)Shi, Min, Yasunaga, Seo, James, Lewis, Zettlemoyer, and tau Yih}]{shi2023replug}
Weijia Shi, Sewon Min, Michihiro Yasunaga, Minjoon Seo, Rich James, Mike Lewis, Luke Zettlemoyer, and Wen tau Yih. 2023.
\newblock \href {http://arxiv.org/abs/2301.12652} {Replug: Retrieval-augmented black-box language models}.

\bibitem[{Tambe et~al.(2021)Tambe, Hooper, Pentecost, Jia, Yang, Donato, Sanh, Whatmough, Rush, Brooks, and Wei}]{tambe2021edgebert}
Thierry Tambe, Coleman Hooper, Lillian Pentecost, Tianyu Jia, En-Yu Yang, Marco Donato, Victor Sanh, Paul Whatmough, Alexander~M. Rush, David Brooks, and Gu-Yeon Wei. 2021.
\newblock \href {https://doi.org/10.1145/3466752.3480095} {Edgebert: Sentence-level energy optimizations for latency-aware multi-task nlp inference}.
\newblock In \emph{MICRO-54: 54th Annual IEEE/ACM International Symposium on Microarchitecture}, MICRO '21, page 830–844, New York, NY, USA. Association for Computing Machinery.

\bibitem[{Touvron et~al.(2023)Touvron, Lavril, Izacard, Martinet, Lachaux, Lacroix, Rozi{\`e}re, Goyal, Hambro, Azhar et~al.}]{touvron2023llama}
Hugo Touvron, Thibaut Lavril, Gautier Izacard, Xavier Martinet, Marie-Anne Lachaux, Timoth{\'e}e Lacroix, Baptiste Rozi{\`e}re, Naman Goyal, Eric Hambro, Faisal Azhar, et~al. 2023.
\newblock Llama: Open and efficient foundation language models.
\newblock \emph{arXiv preprint arXiv:2302.13971}.

\bibitem[{Wang et~al.(2020)Wang, Wang, Huang, Song, and Qin}]{wang2020asurvey}
Bo~Wang, Changhai Wang, Wanwei Huang, Ying Song, and Xiaoyun Qin. 2020.
\newblock \href {https://doi.org/10.1109/ACCESS.2020.3029649} {A survey and taxonomy on task offloading for edge-cloud computing}.
\newblock \emph{IEEE Access}, 8:186080--186101.

\bibitem[{Wu et~al.(2016)Wu, Schuster, Chen, Le, Norouzi, Macherey, Krikun, Cao, Gao, Macherey, Klingner, Shah, Johnson, Liu, Łukasz Kaiser, Gouws, Kato, Kudo, Kazawa, Stevens, Kurian, Patil, Wang, Young, Smith, Riesa, Rudnick, Vinyals, Corrado, Hughes, and Dean}]{wu2016googles}
Yonghui Wu, Mike Schuster, Zhifeng Chen, Quoc~V. Le, Mohammad Norouzi, Wolfgang Macherey, Maxim Krikun, Yuan Cao, Qin Gao, Klaus Macherey, Jeff Klingner, Apurva Shah, Melvin Johnson, Xiaobing Liu, Łukasz Kaiser, Stephan Gouws, Yoshikiyo Kato, Taku Kudo, Hideto Kazawa, Keith Stevens, George Kurian, Nishant Patil, Wei Wang, Cliff Young, Jason Smith, Jason Riesa, Alex Rudnick, Oriol Vinyals, Greg Corrado, Macduff Hughes, and Jeffrey Dean. 2016.
\newblock \href {http://arxiv.org/abs/1609.08144} {Google's neural machine translation system: Bridging the gap between human and machine translation}.

\bibitem[{Yujian and Bo(2007)}]{yujian2007normalized}
Li~Yujian and Liu Bo. 2007.
\newblock A normalized levenshtein distance metric.
\newblock \emph{IEEE transactions on pattern analysis and machine intelligence}, 29(6):1091--1095.

\bibitem[{Zhang et~al.(2022)Zhang, Roller, Goyal, Artetxe, Chen, Chen, Dewan, Diab, Li, Lin et~al.}]{zhang2022opt}
Susan Zhang, Stephen Roller, Naman Goyal, Mikel Artetxe, Moya Chen, Shuohui Chen, Christopher Dewan, Mona Diab, Xian Li, Xi~Victoria Lin, et~al. 2022.
\newblock Opt: Open pre-trained transformer language models.
\newblock \emph{arXiv preprint arXiv:2205.01068}.

\bibitem[{Zhang et~al.(2020)Zhang, Kishore, Wu, Weinberger, and Artzi}]{bert-score}
Tianyi Zhang, Varsha Kishore, Felix Wu, Kilian~Q. Weinberger, and Yoav Artzi. 2020.
\newblock \href {https://openreview.net/forum?id=SkeHuCVFDr} {Bertscore: Evaluating text generation with bert}.
\newblock In \emph{International Conference on Learning Representations}.

\end{thebibliography}
